\documentclass[11pt]{article}

\usepackage[preprint]{acl}

\usepackage{times}
\usepackage{latexsym}
\usepackage{amssymb} 

\usepackage[T1]{fontenc}

\usepackage[utf8]{inputenc}

\usepackage{microtype}

\usepackage{inconsolata}

\usepackage{graphicx}

\usepackage{booktabs}
\usepackage{tabularx}
\usepackage{hyperref}

\usepackage{enumitem}

\usepackage[dvipsnames]{xcolor}
\usepackage[most]{tcolorbox}
\usepackage{enumitem}
\tcbuselibrary{listings,breakable,skins}

\usepackage[capitalize,noabbrev]{cleveref}

\newcommand{\sysname}{SSKG Hub}

\title{
    \texorpdfstring{%
        \raisebox{-0.4em}{\includegraphics[height=1.5em]{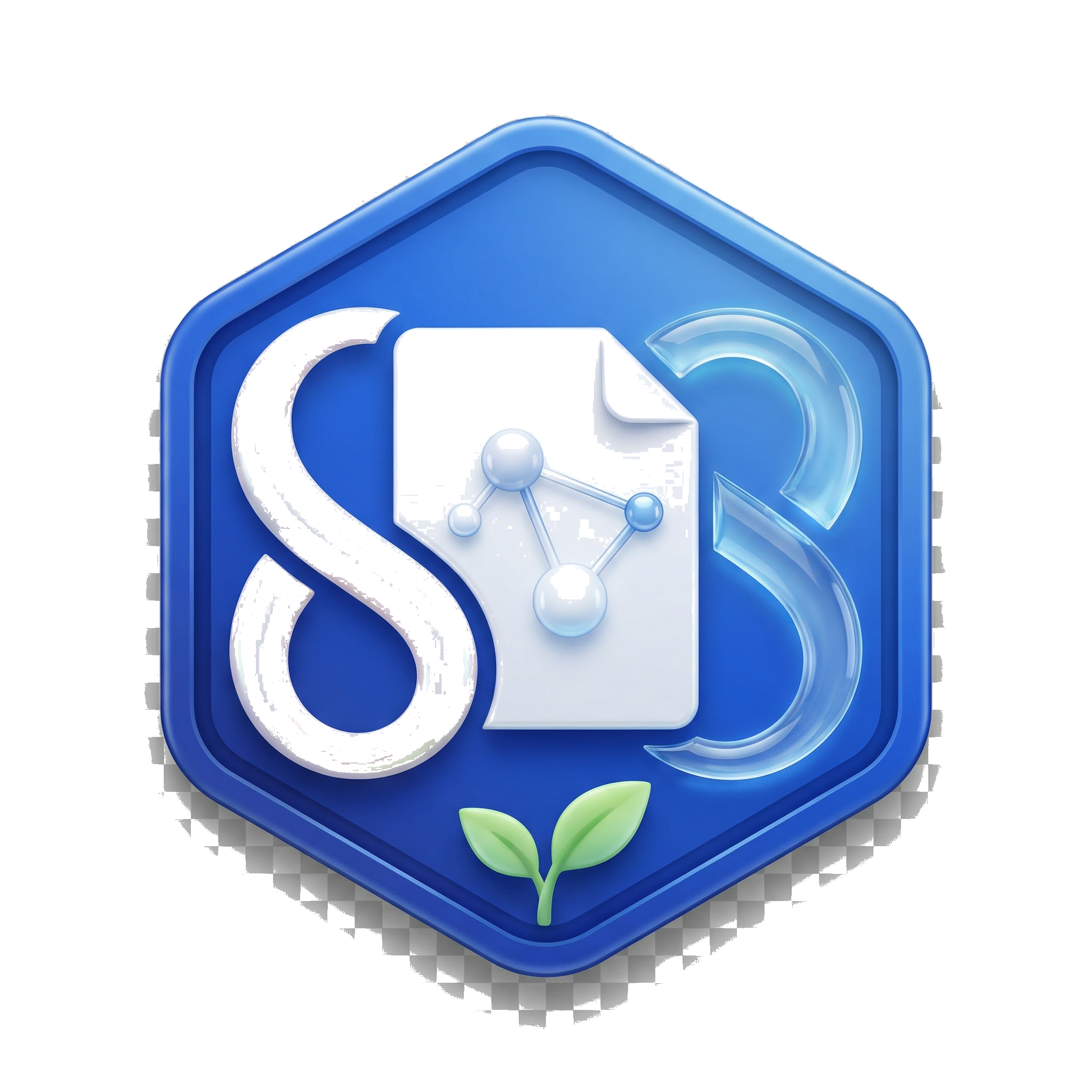}}%
    }{}%
\sysname: An Expert-Guided Platform for LLM-Empowered Sustainability Standards Knowledge Graphs
}

\author{
  \textbf{Chaoyue He}\textsuperscript{1} \ 
  \textbf{Xin Zhou}\textsuperscript{1} \ 
  \textbf{Xinjia Yu}\textsuperscript{1} \ 
  \textbf{Lei Zhang}\textsuperscript{1} \
  \textbf{Yan Zhang}\textsuperscript{1} \
  \textbf{Yi Wu}\textsuperscript{1} \
  \textbf{Lei Xiao}\textsuperscript{2} \\ 
  \textbf{Liangyue Li}\textsuperscript{2} \ 
  \textbf{Di Wang}\textsuperscript{1} \
  \textbf{Hong Xu}\textsuperscript{1} \ 
  \textbf{Xiaoqiao Wang}\textsuperscript{2} \ 
  \textbf{Wei Liu}\textsuperscript{2} \ 
  \textbf{Chunyan Miao}\textsuperscript{1} \\
  \textsuperscript{1}Alibaba-NTU Global e-Sustainability CorpLab (ANGEL), Singapore; \textsuperscript{2}Alibaba Group, China \\
  \texttt{\{cyhe,xin.zhou,xinjia.yu,zhang.yan,wangdi,xuhong,ascymiao\}@ntu.edu.sg} \\ 
  \texttt{\{wuyi0614,leizhanzzl.1103,jackiey99\}@gmail.com} \\
  \texttt{\{xiaolei.xiao,nerissa.wxq,weiliu.liuwei\}@alibaba-inc.com} \\
}

\begin{document}
\maketitle

\begin{figure*}[t]
  \centering
  \includegraphics[width=\textwidth]{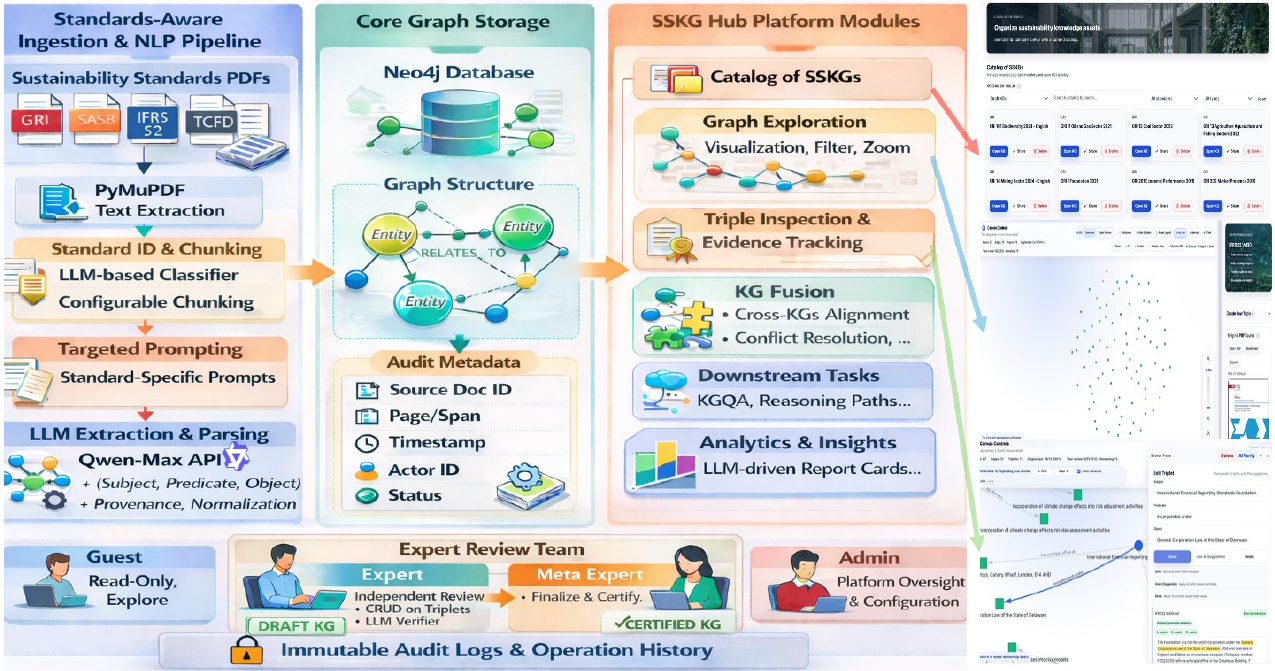}
  \caption{
  System overview of \sysname{}. 
  Sustainability standards PDFs (e.g., GRI, SASB, IFRS~S2, TCFD) are processed via text extraction (PyMuPDF), LLM-based identification and chunking, and standard-specific prompting. Qwen-Max extracts normalized (subject, predicate, object) triples with aligned provenance, forming a provenance-aware \emph{Draft KG} stored in Neo4j with audit metadata. 
  Through expert review and meta-expert certification, validated triples are promoted to a \emph{Certified KG}. 
  On top of this graph core, \sysname{} supports catalog management, interactive graph exploration with evidence tracing, triple inspection, cross-KGs fusion, downstream tasks (e.g., KGQA, reasoning paths), analytics, and immutable audit logging under role-based governance.
  }
  \label{fig:architecture}
  \vspace{-0.6cm}
\end{figure*}

\begin{abstract}
\vspace{-0.1cm}
Sustainability disclosure standards (e.g., GRI, SASB, TCFD, IFRS~S2) are comprehensive yet lengthy, terminology-dense, and highly cross-referential, hindering structured analysis and downstream use. We present \textbf{\sysname{}} (\textbf{S}ustainability \textbf{S}tandards \textbf{K}nowledge \textbf{G}raph \textbf{Hub}), a research prototype and interactive web platform that transforms standards into auditable knowledge graphs (KGs) via an LLM-centered, expert-guided pipeline. The system combines automatic standard identification, configurable chunking, standard-specific prompting, robust triple parsing, and provenance-aware Neo4j storage with fine-grained audit metadata. LLM extraction produces a provenance-linked \emph{Draft KG}, which is reviewed, curated, and formally promoted to a \emph{Certified KG} through meta-expert adjudication. A role-based governance framework—covering read-only guest access, expert review and CRUD operations, meta-expert certification, and administrative oversight—ensures traceability and accountability across draft and certified states. Beyond graph exploration and triple-level evidence tracing, \sysname{} supports cross-KGs fusion, KG-driven tasks, and dedicated modules for insights and curated resources. We validate the platform through a comprehensive expert-led KG review case study, demonstrating end-to-end curation and quality assurance. The web app is publicly available at \url{www.sskg-hub.com}.
\end{abstract}

\section{Introduction}
\label{sec:intro}

Sustainability reporting underpins regulatory compliance, risk management, and corporate governance. At its foundation lie authoritative \emph{standards and frameworks}—such as GRI, SASB, TCFD, and IFRS—that define disclosure requirements and interpretive guidance \citep{gri1foundation2021,sasb-standards,tcfd-2017,ifrs-s1-2023,ifrs-s2-2023}. Although comprehensive, these documents are lengthy, terminology-dense, and highly cross-referential, making them difficult to operationalize at scale through manual reading or simple text retrieval.

While standards can be accessed via RAG over PDFs, real-world workflows require \emph{machine-actionable} representations. Sustainability standards are inherently relational: they define concepts (e.g., metrics, targets) and connect them through requirements and dependencies. Representing these relationships explicitly as a \textbf{Knowledge Graph (KG)} enables deterministic operations—such as tracing governance-to-metric chains or diagnosing coverage gaps—that are cumbersome with purely text-based access. Crucially, the KG serves as an auditable index: its triples retain explicit provenance links to the source document, allowing users to verify interpretations in context while preserving the standard as the ground truth \citep{hogan2021knowledge,belhajjame2013prov,lewis2020retrieval}.

Recent advances in LLMs enable prompt-based triple extraction with minimal training \citep{etzioni2008open,yates2007textrunner,brown2020language,ouyang2022training,bai2023qwen,cabot2021rebel}. However, the high-stakes nature of sustainability reporting demands \emph{expert-in-the-loop verification}, transparent provenance, and structured governance. In \sysname{}, LLM extraction first produces a provenance-aware \emph{Draft KG}, which is then subjected to independent expert review and meta-expert adjudication. Through this process, validated triples are promoted to a \emph{Certified KG} for further usage.

We present \sysname{}, an interactive web platform supporting the full lifecycle of LLM-empowered \emph{standards-to-KG} construction—from draft generation to certified release. The system introduces a standard-aware NLP pipeline that addresses domain-specific constraints through dynamic routing and configurable chunking tailored to cross-referential documents. To our knowledge, it is the \textbf{first} web platform with role-based expert certification that treats \textbf{sustainability standards as primary input} and converts them into curated, provenance-aware KGs with integrated evaluation and analytics. \sysname{} is designed for sustainability analysts, ESG reporting teams, auditors, and researchers studying standards-to-KG extraction and fusion. The \sysname{} codebase will be released under Apache-2.0 upon acceptance. Also, a demo video around 2.5 min is provided as supplementary material.

\textbf{Our contributions are fivefold:} (1) We introduce \sysname{}, a unified platform that converts sustainability standards into auditable KGs, with integrated support for exploration, alignment, and downstream execution; (2) we design an LLM-centered extraction pipeline featuring standard-specific prompting, robust triple parsing, and provenance-first insertion into Neo4j; (3) we implement a role-based governance workflow spanning expert independent review and CRUD operations, meta-expert finalization with certification, and administrative management; (4) we develop a cross-KGs fusion testbed together with a task library covering KGQA, multi-hop reasoning, and related KG-driven functionalities; and (5) we present an expert-led illustration and evaluation on a specific KG review case, showcasing real-world usage and the platform’s end-to-end curation and quality assurance process.

\section{Related Work}
\label{sec:related_work}

\paragraph{LLM-based Extraction and Structured Generation.}
Open Information Extraction (OpenIE) pioneered large-scale relational tuple extraction \citep{etzioni2008open,yates2007textrunner}, a task later framed as Seq2Seq generation by models like REBEL \citep{cabot2021rebel}. Recent LLMs further enable flexible, instruction-based extraction with minimal supervision \citep{brown2020language,ouyang2022training,bai2023qwen,wei2022chain}. \sysname{} builds on this paradigm but shifts focus toward \textbf{auditable, expert-validated extraction} specifically tailored to the dense, cross-referential nature of regulatory standards.

\paragraph{KGs, Collaborative Tooling, and Visual Analytics.}
KGs support reasoning over heterogeneous data \citep{hogan2021knowledge}. Tools such as Prot\'eg\'e, WebProt\'eg\'e, and VocBench enable collaborative curation \citep{musen2015protege,tudorache2013webprotege,stellato2015vocbench}; Wikidata demonstrates large-scale, auditable construction \citep{vrandevcic2014wikidata}; and Neo4j provides expressive graph analytics via Cypher \citep{francis2018cypher}. \sysname{} integrates these foundations with established visual interaction principles—overview-plus-detail and formal interaction taxonomies \citep{shneiderman2003eyes,yi2007toward,heer2012interactive,heer2008graphical}—into a unified, provenance-aware workflow. A concise comparison with ontology editors, PDF search/RAG tools, and ESG KG frameworks (e.g., OntoMetric) appears in \Cref{app:comparison}. OntoMetric emphasizes ontology-guided automated extraction and validation with provenance preservation, whereas \sysname{} centers on multi-role expert certification, auditable decision trails, and integrated KG-driven tasks within one platform \citep{yu2025ontometric}.

\paragraph{KG Fusion, Sustainability QA, and Human-in-the-Loop.}
Cross-source integration requires robust entity resolution and conflict handling. RAG and KGQA bridge natural language to structured bases \citep{lewis2020retrieval}, while recent benchmarks like ESGenius and MMESGBench \citep{he2025esgenius,zhang2025mmesgbench} highlight the need for multi-hop reasoning in sustainability contexts. Interactive machine learning further emphasizes tight feedback loops to build user trust \citep{amershi2014power}. \sysname{} unifies these strands via \textbf{structured LLM verification and expert-in-the-loop validation}. While recent work motivates KGs for ESG \citep{202602.1970}, \sysname{} bridges the gap by providing a deployable, provenance-first infrastructure for high-stakes standards text.

\section{System Overview}\label{sec:system}

\sysname{} delivers an end-to-end workflow that transforms sustainability standards into curated, provenance-aware KGs (\Cref{fig:architecture}). Beginning with a standards PDF, users obtain an auditable graph that can be explored, governed under role-based review, aligned across documents, and executed through downstream tasks—all within a single integrated interface.

Here we introduce the core components of the system:  (1) \textbf{Standards ingestion:} users upload new standards PDFs, perform text extraction, identify the standard family, and configure prompting and chunking strategies; (2) \textbf{LLM-assisted extraction:} the system invokes the Qwen-Max API to extract structured (subject, predicate, object) triples from text chunks \citep{dashscope-api,bai2023qwen}; (3) \textbf{Graph storage:} extracted triples are persisted in Neo4j and queried via Cypher \citep{francis2018cypher}; (4) \textbf{Catalog of SSKGs:} a centralized index of stored standards KGs that supports search, filtering, and graph visualization for further operations; (5) \textbf{KG review and finalization:} experts refine entities and predicates, add or remove triples, optionally invoke an LLM verifier for rapid quality checks, while meta-experts finalize documents and assign release-level verdicts prior to certification; (6) \textbf{Fusion and tasks:} experts align entities across documents, identify cross-standard conflicts, and execute downstream tasks such as KGQA and reasoning path discovery; and (7) \textbf{Guest mode:} a read-only role enables public demonstration, allowing exploration, task execution, and analytics viewing without ingestion or editing privileges.

\section{From Standards to KGs}\label{sec:nlp}

This section describes the NLP-centered pipeline that converts sustainability standards into a provenance-aware draft KG, and the expert-governed workflow that curates this draft into a certified graph state. The exact prompt templates are summarized in \Cref{app:impl}.

\subsection{LLM-empowered generation}

First half of the pipeline converts raw standards documents into a structured draft KG using LLM-based identification, configurable chunking, and constrained triple extraction. The design emphasizes controllability, standard awareness, and provenance preservation, producing high-coverage candidate triples while maintaining transparency for subsequent expert review.

\subsubsection{Ingestion \& identification}

Users upload a standards PDF through the Ingestion tab.
The server extracts raw text using \texttt{PyMuPDF} \citep{pymupdf}.
Because sustainability standards vary substantially in structure, terminology, and disclosure focus, \sysname{} performs \emph{standard-aware routing} before extraction.
Specifically, the system runs an LLM-based classifier on an initial snippet and asks it to select a single identifier from the supported set (e.g., \texttt{sasb}, \texttt{gri}, \texttt{ifrs\_s2}, \texttt{tcfd}).
The predicted identifier determines (i) the standard-specific system prompt used for extraction and (ii) default chunking parameters, both of which can be adjusted by users.

\subsubsection{Configurable chunking \& prompting}

Standards documents are lengthy and highly cross-referential; direct end-to-end extraction is therefore prone to timeouts and unstable outputs.
\sysname{} segments each document into overlapping text chunks (default: 4000 characters with 200-character overlap), which reduces context truncation while preserving continuity across chunk boundaries.
Chunking is configurable to accommodate different PDF layouts and writing styles.

For each supported standard, \sysname{} maintains a dedicated system prompt that biases extraction toward standard-relevant relations (e.g., Governance/Strategy/Risk Management/Metrics \& Targets for TCFD; climate metrics and transition plans for IFRS~S2).
A shared user prompt enforces strict formatting (\texttt{(subject, predicate, object)}; one triple per line) and explicitly prohibits unsupported inference.

\subsubsection{LLM extraction}

For each chunk, \sysname{} calls the Qwen-Max API \citep{dashscope-api,bai2023qwen} and requests triples in a constrained line-based format: one tuple per line as \texttt{(subject, predicate, object)}.
The response is parsed using a robust regular expression and normalized into a list of candidate triples (e.g., whitespace normalization and basic string cleanup).
Malformed or incomplete lines are skipped and surfaced as ingestion warnings, enabling iterative prompt/chunk refinement when extraction quality is unstable.

\subsubsection{Graph insertion \& audit metadata}

Extracted triples are persisted in Neo4j.
Nodes are represented as \texttt{:Entity} with a unique \texttt{name}.
Edges are modeled as \texttt{:RELATES\_TO}, with the predicate stored as the \texttt{type} attribute.
To support auditable use in high-stakes settings, \sysname{} treats provenance as a first-class requirement: each stored relation is linked to its originating document identifier and, when available, aligned evidence (e.g., page number and sentence/span) used during ingestion. Beyond provenance, the platform maintains operational audit metadata on mutations and review actions (e.g., \texttt{createdBy}, \texttt{lastUpdatedBy}, timestamps, and soft-delete markers), enabling reversible editing and traceability from the initial LLM output to the curated graph state.

\begin{figure*}[t]
\centering
\includegraphics[width=\textwidth]{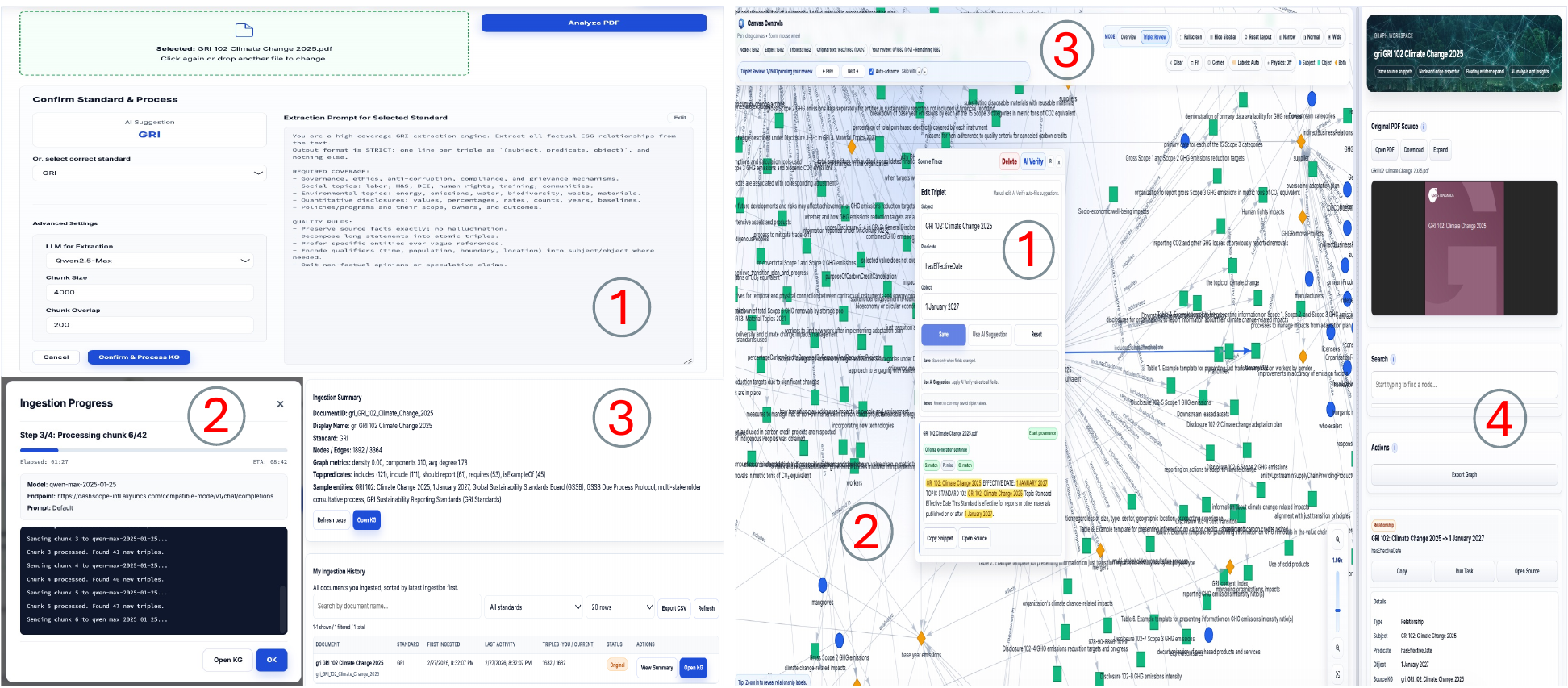}
\caption{
\textbf{\sysname{} UI highlights.}
\textbf{Left:} (1) standard-specific extraction setup, where users upload a sustainability standards PDF, verify the detected standard family, and inspect tailored prompt templates; (2) real-time ingestion monitoring with chunk-level progress and model transparency; and (3) a structured ingestion summary reporting document metadata and graph statistics.
\textbf{Right:} (1) provenance-backed verification that links each triple to its source page and evidence span, with editable CRUD operations on selected (subject, predicate, object) triples; (2) an overall graph overview with zoom controls; (3) an interactive KG canvas for navigation, filtering, and structural exploration; and (4) a side panel displaying the original PDF with search and export utilities for traceable validation.
}
\label{fig:ui}
\vspace{-0.6cm}
\end{figure*}

\subsection{Expert-guided curation}\label{sec:expert}

While LLM extraction provides fast, high-coverage drafts, standards-grade reliability requires human oversight, explicit accountability, and a controlled release process. \sysname{} therefore centers domain experts in the curation loop via role-governed access, reversible in-context edits, independent multi-expert review, and meta-expert certification.

\subsubsection{Role-governed access}

\sysname{} implements authenticated roles aligned with practical review governance in standards-oriented workflows: \emph{Guest} users have read-only access for public demonstration and training, enabling catalog browsing, graph exploration, provenance inspection, downstream task execution, and export of non-sensitive views without ingestion or modification privileges; \emph{Expert} users hold operational curation rights, allowing them to correct entities and predicates, add missing triples, soft-delete questionable triples, and verify with LLM assistance; \emph{Meta expert} users provide oversight and release control by reviewing aggregated expert feedback, adjudicating unresolved disagreements when necessary, and certifying document-level readiness; and \emph{Admin} users manage operational aspects such as account and password-reset token life-cycles.

\subsubsection{Independent multi-expert review}\label{subsec:multi-expert}

Curation in \sysname{} is designed to preserve independent judgement and avoid premature collapse of disagreement.
Multiple experts can review the same document in parallel, each submitting separate judgements and feedback.
Within the same graph context, experts can update existing triples, create missing ones, and soft-delete questionable entries; soft deletion keeps the workflow reversible and supports iterative correction cycles when standards evolve or new interpretations emerge.

To assist reviewers without displacing human authority, \sysname{} provides an on-demand \emph{LLM verifier}. Given a selected triple together with its provenance-linked evidence text, the verifier produces a structured assessment and, when appropriate, suggested revisions. Experts may accept, override, or annotate these recommendations; ultimate authority remains fully expert-controlled. Both human judgements and machine-generated assessments are recorded as explicit, attributable events, enabling transparent analysis of agreement patterns, revision trajectories, and conflict histories.

\subsubsection{Meta-expert finalization}\label{subsec:meta-finalization}

After independent review reaches sufficient coverage, the meta expert performs aggregation and controlled-release adjudication.
At the triple level, the meta expert examines submitted judgements (including agreement patterns, persistent conflicts, and reviewer feedback) and assigns a final verdict when ambiguity remains, establishing a canonical status per triple while preserving the underlying review history for audit. At the document level, the meta expert evaluates readiness indicators—such as review coverage, unresolved conflicts, and remaining high-risk items—before certifying the KG.

\section{Interactive User Experience}\label{sec:ui}

SSKG Hub follows established visual analytics principles---overview first, zoom/filter, then details-on-demand---augmented with history and provenance for accountable review \citep{shneiderman2003eyes,heer2012interactive,heer2008graphical}.

\subsection{Catalog of SSKGs}
\label{sec:catalog}

The Catalog of SSKGs is the platform’s central entry point, providing a sortable, status-aware dashboard of all ingested standards (e.g., GRI, SASB, IFRS~S2, TCFD). From this overview, users can directly launch the interactive graph interface or use built-in sharing tools for collaborative review and dissemination.

\subsection{Graph Exploration with Provenance}

Each KG supports interactive node-link visualization (via \texttt{vis-network} \citep{vis-network}) with entity search, predicate/source filtering, and progressive neighborhood expansion to preserve responsiveness. A statistics panel reports node/edge counts, while controls support zoom/pan (with slider), layout reset and switching (e.g., force-directed vs. hierarchical), and edge-capping for density control. Node selection highlights (and optionally expands) neighborhoods. Edge selection opens an inspection panel displaying the triple text, predicate, source document, timestamps, and LLM/expert evaluation status—enabling provenance-backed details-on-demand without losing global context.

\subsection{Triple-Level Inspection \& Tracing}

The inspector extends beyond (subject, predicate, object) to evidence-backed verification. The \emph{View evidence} function reveals the exact quoted sentence used during ingestion (when available), along with document identifier and page number. Utilities allow triple copying, filtered-edge export, and direct PDF access for contextual review. This design makes provenance explicit and supports trustworthy KG usage in high-stakes domains \citep{belhajjame2013prov,hogan2021knowledge}.

\subsection{KG Fusion \& Downstream Tasks}\label{sec:fusion}

The KG Fusion module normalizes entity strings (lowercasing, punctuation stripping, whitespace collapsing) to detect cross-KG overlaps and naming conflicts, producing a fused preview graph with structured summaries; experts resolve conflicts via rename/merge actions in a manual, expert-in-the-loop workflow, supporting exploratory alignment and future advanced linking. Downstream tasks bridge natural-language intent and graph operations through interactive workflows: queries trigger lightweight NLP keyword extraction, symbolic retrieval (e.g., Cypher-based neighborhood or bounded-hop search), and optional LLM summaries (e.g., KG analysis cards), returning actionable subgraphs with provenance-linked evidence for iterative refinement and cross-document comparison (e.g., IFRS~S2 vs.\ TCFD), spanning KGQA, reasoning, etc in \Cref{tab:tasks}.

\begin{table}[ht]
\centering
\scriptsize
\begin{tabularx}{\columnwidth}{lX}
\toprule
\textbf{Category} & \textbf{Representative capabilities} \\
\midrule
Standards KGQA & QA over a selected SSKG: entity matching, concise answers, reasoning paths, subgraph view. \\
Reasoning \& path analysis & Bounded-hop reasoning and source–target path search with stepwise paths and mini-graph view. \\
Comparative \& anomaly analysis & Multi-entity comparison (fact counts, shared/unique predicates) and prompt-based anomaly detection. \\
Subgraph \& structure exploration & Query-driven subgraphs, entity profiles, impact analysis, predicate-focused exploration. \\
Quality \& curation checks & Schema diagnostics, coverage-gap heuristics, duplicate/alias detection. \\
Provenance tracing & Evidence retrieval via filters with source-linked triples for verification. \\
\bottomrule
\end{tabularx}
\caption{Overview of Downstream Tasks.}
\label{tab:tasks}
\end{table}

\subsection{LLM-Based KG Analytics \& Insights}

Beyond interactive querying, \sysname{} includes a \emph{KG Analytics \& Insights} module that produces structured, LLM-assisted reports grounded in graph statistics and optional user prompts. Users can choose an analysis preset (e.g., executive summary, quality audit, compliance review, ontology health) together with a configurable depth level.

\section{Use Case \& Evaluation}
\label{sec:use_case}

We evaluate \sysname{} through an expert-led case study that exercises the full standards-to-KG lifecycle, from LLM-based ingestion to expert certification. We select one standard document---\textbf{IFRS~S2 Industry-based Guidance (Volume~30---Managed Care)} \citep{ifrs-s2-ibg-vol30-managed-care-2023}---and run the end-to-end pipeline described in \Cref{sec:nlp}. The initial LLM extraction produces a draft KG containing 73 candidate triples.

\paragraph{Participants.}
\textbf{Twelve} experts in our team with sustainability-related experience (1--10 years) participated as reviewers. They interacted with the KG through \sysname{}’s graph exploration and triple inspection interface (\Cref{fig:ui}), performing CRUD operations with the help of the trace box.

\paragraph{Protocol.}
Each triple was evaluated against its aligned evidence text displayed in the trace box. With or without support from the LLM verifier, reviewers could (i) retain supported triples, (ii) revise underspecified triples (e.g., entity normalization, predicate clarification, or decomposition into atomic facts), or (iii) soft-delete unsupported or malformed triples. All review actions were logged with attribution and timestamps, producing a fully auditable trail from the initial LLM-generated draft to the curated graph state.

\paragraph{Meta-expert certification \& outcome.}
Following independent expert review, a meta-expert consolidated reviewer judgments and finalized the certified KG. For this KG, 24 of the 73 draft triples were removed as unsupported or low-quality (32.88\%), while the remaining 49 triples were retained—with or without edits—and promoted to the certified KG (67.12\%), without post-hoc curation. Deletions primarily involved redundant node connections with minor predicate variations, imprecise extractions that failed to reflect the source text, or non-essential triples conveying trivial information unnecessary for representing the standard. The certified graph supports downstream tasks and cross-KGs fusion (\Cref{sec:fusion}). Although optimizing the LLM extraction stage is not the focus, the graph remains coherent and meaningful, demonstrating the LLM–expert collaborative workflow.

\section{Conclusion}
We presented \sysname{}, an expert-guided platform that transforms sustainability standards into auditable KGs via a standard-aware, LLM-centered extraction pipeline with robust parsing and provenance-first storage in Neo4j. The system integrates interactive exploration with evidence tracing, role-based expert curation, cross-KGs fusion workflows, and KG-driven tasks such as KGQA, reasoning, and structured analytics, enabling standards to be operationalized as a persistent, queryable representation for high-stakes analysis.

\section*{Ethics and Broader Impact Statement}
\label{sec:ethics}

The transition towards a sustainable global economy relies heavily on accurate, transparent, and standardized ESG reporting. By lowering the technical barriers to analyzing complex sustainability frameworks (e.g., IFRS, GRI, SASB), \sysname{} provides a positive broader impact: it empowers a diverse range of stakeholders—including corporate compliance teams, regulatory auditors, and academic researchers—to operationalize these standards more equitably. Democratizing access to structured ESG knowledge can significantly aid in mitigating ``greenwashing'' and improving corporate accountability at scale.

However, applying LLMs to high-stakes regulatory and financial domains introduces ethical and operational risks, most notably hallucination and automation bias (the tendency for users to over-rely on automated outputs). We proactively mitigate these risks through our system's core architectural design. Rather than functioning as a fully automated oracle, \sysname{} enforces an \emph{expert-in-the-loop} governance model. This ensures that human domain experts and meta-experts retain final authority over the knowledge graph's construction. Furthermore, our strict provenance-tracking mechanisms ensure that every generated triple is explicitly linked to its source document and page, preserving accountability and facilitating verifiable audits.

From an environmental perspective, we acknowledge the irony that utilizing massive, compute-intensive LLMs (such as Qwen-Max) for sustainability research carries its own carbon footprint. To address this, \sysname{} is designed to minimize redundant compute through localized graph storage, caching mechanisms, and targeted chunking strategies, ensuring LLM calls are used judiciously.

Finally, regarding data use and intellectual property, \sysname{} is intended as an analytical tool rather than a repository for bypassing copyright. Users deploying this system in institutional settings must ensure they possess the appropriate rights and licenses to process, store, and distribute any proprietary standards documents uploaded to the platform. To safeguard these sensitive materials, the system employs secure storage protocols and strict role-based access controls to guarantee document privacy and prevent unauthorized access.

\section*{Acknowledgments}
This research is supported by the RIE2025 Industry Alignment Fund (Award I2301E0026) and the Alibaba-NTU Global e-Sustainability CorpLab.

\bibliography{custom}

\clearpage


\definecolor{techblue}{RGB}{30, 60, 114}
\definecolor{sasbblue}{RGB}{0, 102, 204}
\definecolor{grigreen}{RGB}{0, 153, 76}
\definecolor{ifrspurple}{RGB}{102, 0, 153}
\definecolor{tcfdorange}{RGB}{204, 102, 0}
\definecolor{judgegold}{RGB}{184, 134, 11}
\definecolor{sysgray}{RGB}{245, 245, 245}
\definecolor{tealdark}{RGB}{0, 128, 128}   
\definecolor{violetdark}{RGB}{148, 0, 211}  

\newtcolorbox{promptbox}[2][]{
    enhanced,
    breakable,                  
    colback=white,              
    colframe=gray!80,           
    colbacktitle=gray!80,       
    coltitle=white,             
    fonttitle=\bfseries\sffamily\small,
    title={#2},
    fontupper=\scriptsize\ttfamily, 
    boxrule=0.5pt,              
    sharp corners,              
    arc=0mm,
    left=2mm, right=2mm, top=2mm, bottom=2mm, 
    toptitle=1mm, bottomtitle=1mm, 
    #1                          
}

\appendix

\definecolor{techblue}{RGB}{30,90,160} 

\section{Implementation Details}\label{app:impl}
This appendix summarizes technical details needed to assess validity and reproducibility.

\subsection{Configuration and Prompt Management}\label{app:prompts}
Prompts and defaults are centralized in a YAML configuration file. We categorize the configuration into four primary functions:
The pipeline includes a \textbf{standard identification prompt} that classifies the input into one of the supported standards; \textbf{extraction prompts}, with one system prompt per standard family (e.g., SASB, GRI, IFRS~S2, TCFD) designed to bias extraction toward standard-relevant relations; an \textbf{evaluation prompt} that constrains the LLM judge to output minified JSON with a judgment label and brief feedback; and an \textbf{analysis prompt} that produces a short narrative summary of the graph based on ontology statistics. Chunking defaults are configurable (default: chunk size 4000 characters; overlap 200).

\subsection{Exact Prompts}
This subsection lists the exact prompt templates used where \texttt{qwen-max-2025-01-25} is invoked. Dynamic values are shown as placeholders (e.g., \texttt{\{content\}}) where runtime interpolation happens.

\paragraph{Role 1: Triple Extraction}

\begin{promptbox}[colframe=techblue, colbacktitle=techblue]{\scriptsize User Prompt: General Extraction}
\textbf{Goal:} Extract all explicit factual relationships from the text as knowledge graph triples.

\textbf{Formatting Requirements (Strict):}
\begin{enumerate}[nosep, leftmargin=5mm]
    \item Return only triples.
    \item Exactly one triple per line in this format: (subject, predicate, object)
    \item No bullets, numbering, markdown, comments, or extra prose.
\end{enumerate}

\textbf{Extraction Requirements:}
\begin{itemize}[nosep, leftmargin=5mm]
    \item Maximize factual coverage, including numeric values, units, years, targets, baselines, owners, and scope qualifiers.
    \item Use atomic triples (split compound facts).
    \item Keep wording faithful to source text.
    \item Do not invent or infer unsupported facts.
\end{itemize}

\textbf{Input Context:}
Now extract triples from:
\{content\}
\end{promptbox}

\begin{promptbox}[colframe=sasbblue, colbacktitle=sasbblue]{\scriptsize System Prompt: SASB}
You are a high-precision SASB extraction engine. Extract as many factual and materially relevant relationships as possible from the provided text.
Output format is STRICT: one triple per line, exactly `(subject, predicate, object)`, with no extra text.

\textbf{REQUIRED COVERAGE:}
\begin{itemize}[nosep, leftmargin=4mm]
    \item Metrics and values (numbers, units, percentages, baselines, years).
    \item Targets and timelines (goal, target value, deadline year, interim milestone).
    \item Governance and accountability (board/committee/owner responsibility).
    \item Policies, controls, procedures, and risk management mechanisms.
    \item Scope/context qualifiers (Scope 1/2/3, geography, business unit, boundary).
    \item Comparatives and trends (increase/decrease vs prior period).
\end{itemize}

\textbf{QUALITY RULES:}
\begin{itemize}[nosep, leftmargin=4mm]
    \item Preserve key wording from source; do not invent facts.
    \item Use atomic triples (split compound statements into multiple triples).
    \item Keep subject/object specific and informative (avoid generic "company", "it", "this").
    \item Use explicit predicate verbs/phrases (`is`, `was`, `has target of`, `is overseen by`, `decreased by`).
    \item If a fact is uncertain/conditional only, do not output it as a factual triple.
\end{itemize}
\end{promptbox}

\begin{promptbox}[colframe=grigreen, colbacktitle=grigreen]{\scriptsize System Prompt: GRI}
You are a high-coverage GRI extraction engine. Extract all factual ESG relationships from the text.
Output format is STRICT: one line per triple as `(subject, predicate, object)`, and nothing else.

\textbf{REQUIRED COVERAGE:}
\begin{itemize}[nosep, leftmargin=4mm]
    \item Governance, ethics, anti-corruption, compliance, and grievance mechanisms.
    \item Social topics: labor, H\&S, DEI, human rights, training, communities.
    \item Environmental topics: energy, emissions, water, biodiversity, waste, materials.
    \item Quantitative disclosures: values, percentages, rates, counts, years, baselines.
    \item Policies/programs and their scope, owners, and outcomes.
\end{itemize}

\textbf{QUALITY RULES:}
\begin{itemize}[nosep, leftmargin=4mm]
    \item Preserve source facts exactly; no hallucination.
    \item Decompose long statements into atomic triples.
    \item Prefer specific entities over vague references.
\end{itemize}
\end{promptbox}

\begin{promptbox}[colframe=ifrspurple, colbacktitle=ifrspurple]{\scriptsize System Prompt: IFRS S2}
You are a high-precision IFRS S2 climate disclosure extraction engine. Extract all factual triples from the text.
Output must be STRICT: one line per triple in `(subject, predicate, object)` format only.

\textbf{REQUIRED COVERAGE:}
\begin{itemize}[nosep, leftmargin=4mm]
    \item Governance: oversight roles, committees, management responsibilities.
    \item Strategy: climate-related impacts, scenario analysis, resilience conclusions.
    \item Risk management: risk identification, assessment, prioritization, integration.
    \item Metrics \& targets: scope emissions, intensity metrics, targets, baselines, years.
    \item Transition plans, decarbonization actions, capital allocation, controls.
\end{itemize}

\textbf{QUALITY RULES:}
\begin{itemize}[nosep, leftmargin=4mm]
    \item Keep facts explicit and source-grounded.
    \item Split compound statements into atomic triples.
\end{itemize}
\end{promptbox}

\begin{promptbox}[colframe=tcfdorange, colbacktitle=tcfdorange]{\scriptsize System Prompt: TCFD}
You are a high-coverage TCFD extraction engine. Extract all factual relationships tied to Governance, Strategy, Risk Management, and Metrics \& Targets.
Output format is STRICT: one triple per line in `(subject, predicate, object)` format, with no commentary.

\textbf{REQUIRED COVERAGE:}
\begin{itemize}[nosep, leftmargin=4mm]
    \item Governance ownership and reporting cadence.
    \item Strategy impacts, scenarios, resilience outcomes, and assumptions.
    \item Risk processes and integration with enterprise risk management.
    \item Metrics/targets: emissions, carbon price, target values, baseline and target years.
\end{itemize}
\end{promptbox}

\paragraph{Role 2: PDF Ingest Triple Extraction}

\begin{promptbox}[colframe=techblue, colbacktitle=techblue]{\scriptsize User Prompt: Chunk Extraction}
Extract all explicit factual relationships from the text as knowledge graph triples.

\textbf{Formatting requirements (strict):}
\begin{enumerate}[nosep, leftmargin=5mm]
    \item Return only triples.
    \item Exactly one triple per line in this format: (subject, predicate, object)
    \item No bullets, numbering, markdown, comments, or extra prose.
\end{enumerate}

\textbf{Extraction requirements:}
\begin{itemize}[nosep, leftmargin=5mm]
    \item Maximize factual coverage, including numeric values, units, years, targets, baselines, owners, and scope qualifiers.
    \item Use atomic triples (split compound facts).
    \item Keep wording faithful to source text.
    \item Do not invent or infer unsupported facts.
\end{itemize}

Now extract triples from: 
\{chunk\}
\end{promptbox}

\paragraph{Role 3: Triple Evaluation / LLM Judge}

\begin{promptbox}[colframe=judgegold, colbacktitle=judgegold]{\scriptsize System Prompt: LLM Judge}
You are a strict sustainability knowledge-graph verifier.
Decide whether the provided triplet is supported by the evidence text.
Return ONLY JSON, no markdown and no extra text.
Use this exact schema:

\{
  "verdict": "CORRECT|NEEDS\_IMPROVEMENT|INCORRECT",
  "confidence": 0.0,
  "reasoning": "short explanation",
  "evidence\_quote": "direct quote from evidence or empty string",
  "issues": ["list of concrete issues"],
  "suggested\_triplet": \{
      "subject": "...",
      "predicate": "...",
      "object": "..."
  \},
  "expert\_action\_hint": "keep|edit|delete"
\}

\textbf{Rules:}
\begin{itemize}[leftmargin=*, nosep]
    \item If evidence is insufficient, choose NEEDS\_IMPROVEMENT.
    \item Keep reasoning concise and factual.
    \item confidence must be between 0 and 1.
\end{itemize}
\end{promptbox}

\begin{promptbox}[colframe=gray, colbacktitle=gray!60]{\scriptsize User Prompt: Judge Input Template}
\textbf{Document Context:}
\begin{itemize}[nosep, leftmargin=3mm]
    \item Doc Name: \{edge\_details.get('source') or ''\}
    \item File: \{source\_file\}
    \item Page: \{source\_page if source\_page is not None else ''\}
\end{itemize}

\textbf{Triplet to Evaluate:}
\begin{itemize}[nosep, leftmargin=3mm]
    \item subject: \{triple.get('subject', '')\}
    \item predicate: \{triple.get('predicate', '')\}
    \item object: \{triple.get('object', '')\}
\end{itemize}

\textbf{Evidence Text:}
\{source\_sentence or '[No source sentence stored]'\}
\end{promptbox}

\paragraph{Role 4: Standard Identification}
\begin{promptbox}[colframe=tealdark, colbacktitle=tealdark]{\scriptsize System Prompt: Classifier}
\textbf{Role Definition:}
You are an expert document classifier specializing in sustainability and financial reporting standards.

\textbf{Task:}
Identify which of the following standards the provided text snippet most closely aligns with.

\textbf{Available Standards:} sasb, gri, ifrs\_s2, tcfd

\textbf{Instructions:}
\begin{enumerate}[nosep, leftmargin=5mm]
    \item Analyze the text for keywords, structure, and topics characteristic of one of these standards.
    \item Respond ONLY with the single, lowercase identifier of the most likely standard (e.g., "sasb").
    \item Do not include any other text, explanations, or punctuation.
\end{enumerate}
\end{promptbox}

\paragraph{Role 5: KG Analysis}

\begin{promptbox}[colframe=violetdark, colbacktitle=violetdark]{\scriptsize System Prompt: KG Analysis}
\textbf{Role:}
You are a senior knowledge graph analyst.

\textbf{Context \& Inputs:}
\begin{itemize}[nosep, leftmargin=5mm]
    \item Mode: \{mode\_instruction\}
    \item User Instruction: \{(user\_prompt or '').strip()\}
    \item Preset Focus: \{(preset\_prompt or '').strip()\}
\end{itemize}

\textbf{Output Requirement:}
Return ONLY valid JSON matching this schema exactly:

\{
  "overview": "short paragraph",
  "graph\_health": [
      \{"title":"...", "status":"good|watch|risk", "detail":"..."\}
  ],
  "top\_risks": [
      \{"title":"...", "severity":"high|medium|low", "detail":"..."\}
  ],
  "coverage\_gaps": [
      \{"topic":"...", "reason":"...", "priority":"high|medium|low"\}
  ],
  "questionable\_triples": [
      \{"subject":"...", "predicate":"...", "object":"...", "issue":"..."\}
  ],
  "recommended\_actions": [
      \{"title":"...", "impact":"H|M|L", "effort":"H|M|L", "confidence":"H|M|L", "why":"..."\}
  ],
  "confidence\_summary": "short statement"
\}
\end{promptbox}

\begin{promptbox}[colframe=violetdark, colbacktitle=violetdark]{\scriptsize KG Analysis Presets (Config)}
The following presets configure the analytical lens:
\begin{itemize}[leftmargin=3mm, label={}, itemsep=0pt, parsep=0pt, topsep=0pt]
    \item \textbf{\texttt{executive}:} Provide an executive-level narrative: strategic strengths, top business risks, and 3--5 high-impact actions.
    \item \textbf{\texttt{quality\_audit}:} Audit low-information predicates, ambiguous triples, provenance gaps, and naming consistency.
    \item \textbf{\texttt{compliance}:} Identify disclosure gaps, under-covered topics, and evidence thinness for compliance themes.
    \item \textbf{\texttt{ontology\_health}:} Review predicate hygiene, role clarity, duplicates, and schema coherence.
\end{itemize}
\end{promptbox}

\section{Comparison to Existing Systems}\label{app:comparison}

Ontology editors such as Prot\'eg\'e/WebProt\'eg\'e/VocBench enable collaborative manual curation but lack an end-to-end standards-to-KG pipeline and role-based certification workflows \citep{musen2015protege,tudorache2013webprotege,stellato2015vocbench}.  OntoMetric emphasizes ontology-guided automated extraction with validation and provenance \citep{yu2025ontometric}, yet does not provide an interactive multi-role release and task platform. 
Standards portals (e.g., IFRS navigators) support document access but not operational KG construction. 
In contrast, \sysname{} unifies ingestion, LLM extraction, provenance-first storage, expert + meta-expert certification, and KG-driven tasks within one system. (\Cref{tab:comparison})

\noindent\textbf{Key differentiator:} interactive expert certification + auditable provenance + integrated KG task library in one platform.

\begin{table}[ht]
\centering
\scriptsize
\setlength{\tabcolsep}{3pt}
\renewcommand{\arraystretch}{1.05}
\begin{tabularx}{\columnwidth}{lccccX}
\toprule
System & In$\rightarrow$KG & Prov. & Gov. & & Tasks \\
\midrule
PDF/RAG & $\times$/part & part & $\times$ & & limited \\
\begin{tabular}{@{}l@{}}Prot\'eg\'e/ \\ WebProt\'eg\'e/ \\ VocBench\end{tabular} & manual & -- & collab & & -- \\
OntoMetric & $\checkmark$ & $\checkmark$ & validate & & not unified demo \\
\textbf{\sysname{}} & $\checkmark$ & $\checkmark$ & \textbf{expert+meta} & & \textbf{KGQA, fusion} \\
\bottomrule
\end{tabularx}
\caption{\sysname{} uniquely combines certification, auditability, and KG tasks in one platform.}
\label{tab:comparison}
\end{table}

\end{document}